\documentclass[10pt,twocolumn,letterpaper]{article}

\usepackage[pagenumbers]{cvpr} 

\usepackage{graphicx}
\usepackage{amsmath}
\usepackage{amssymb}
\usepackage{booktabs}

\newcommand{\tabincell}[2]{\begin{tabular}{@{}#1@{}}#2\end{tabular}}
\usepackage{multirow}
\usepackage{colortbl}
\definecolor{tabtitle}{gray}{.8}
\usepackage{bm}
\usepackage{pifont}
\newcommand{\yes}{\text{\ding{51}}}

\usepackage{marvosym}

\usepackage[pagebackref,breaklinks,colorlinks]{hyperref}

\usepackage[capitalize]{cleveref}
\crefname{section}{Sec.}{Secs.}
\Crefname{section}{Section}{Sections}
\Crefname{table}{Table}{Tables}
\crefname{table}{Tab.}{Tabs.}

\def\confName{CVPR}
\def\confYear{2023}

\begin{document}
	
	\title{Visual Prompt Multi-Modal Tracking}
	
	\author{Jiawen Zhu$^1$\protect\footnotemark[2] \quad Simiao Lai$^1$\protect\footnotemark[2] \quad Xin Chen$^1$ \quad Dong Wang$^1$\protect\textsuperscript{\Letter} \quad Huchuan Lu$^{1,2}$  \\
		$^1$Dalian University of Technology, China \quad
		$^2$Peng Cheng Laboratory, China \\
	{\tt\small \{jiawen,laisimiao,chenxin3131\}@mail.dlut.edu.cn, \{wdice,lhchuan\}@dlut.edu.cn}}

\maketitle
\renewcommand{\thefootnote}{\fnsymbol{footnote}}
\footnotetext[2]{Equal contribution.}
\footnotetext{\!\!\!\!\textsuperscript{\Letter}Corresponding author: Dr. Dong Wang.}

\begin{abstract}
Visible-modal object tracking gives rise to a series of downstream multi-modal tracking tributaries.
To inherit the powerful representations of the foundation model, a natural modus operandi for multi-modal tracking is full fine-tuning on the RGB-based parameters.
Albeit effective, this manner is not optimal due to the scarcity of downstream data and poor transferability, etc. 
In this paper, inspired by the recent success of the prompt learning in language models, we develop \textbf{Vi}sual \textbf{P}rompt multi-modal \textbf{T}racking (ViPT), which learns the modal-relevant prompts to adapt the frozen pre-trained foundation model to various downstream multi-modal tracking tasks.
ViPT finds a better way to stimulate the knowledge of the RGB-based model that is pre-trained at scale, meanwhile only introducing a few trainable parameters (less than 1\% of model parameters). 
ViPT outperforms the full fine-tuning paradigm on multiple downstream tracking tasks including RGB+Depth, RGB+Thermal, and RGB+Event tracking. 
Extensive experiments show the potential of visual prompt learning for multi-modal tracking, and ViPT can achieve state-of-the-art performance while satisfying parameter efficiency. Code and models are available at \href{https://github.com/jiawen-zhu/ViPT}{https://github.com/jiawen-zhu/ViPT}. 
\end{abstract}

\section{Introduction}
\label{sec:intro}

RGB-based tracking, a foundation task of visual object tracking, gains from 
large-scale benchmarks~\cite{imagenet,coco,trackingnet,lasot,youtube-vos,got10k}
provided by the community, 
and many excellent works~\cite{siamesefc, siameserpn, atom, dimp, transt} have spurted out over the past decades.
Despite the promising results, object tracking based on pure RGB sequences is still prone to failure in some complex and corner scenarios, \eg, extreme illumination, background clutter, and motion blur.
Therefore, multi-modal tracking is drawing increasing attention due to the ability to achieve more robust tracking by utilizing inter-modal complementarity,
among which RGB+Depth (RGB-D)~\cite{depthtrack, rgbd1k}, RGB+Thermal (RGB-T)~\cite{vtuav, apfnet}, and RGB+Event (RGB-E)~\cite{fenet, stnet} are represented.

\begin{figure}[!t]
	\centering
	\includegraphics[width=0.485\textwidth]{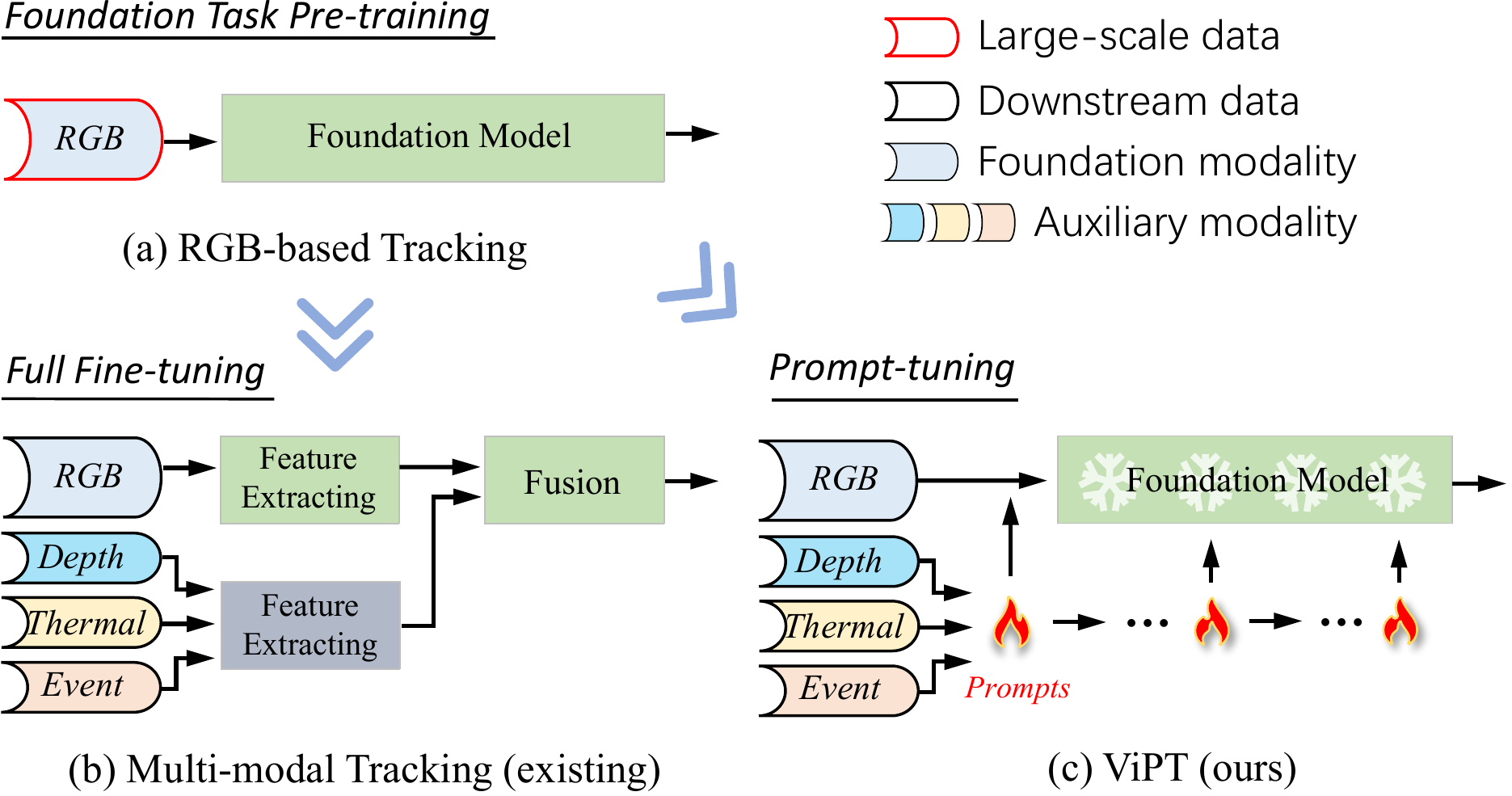}
	\vspace{-6.5mm}
	\caption{\textbf{Existing multi-modal tracking paradigm \emph{vs.} \negthinspace \negthinspace \negthinspace ViPT.} 
		(a) 
		Foundation model training on large-scale RGB sequences.
		(a)$\rightarrow$(b) Existing multi-modal methods extend the off-the-shelf foundation model and 
		conduct full fine-tuning on downstream tracking tasks. (a)$\rightarrow$(c) We investigate the design of a multi-modal tracking method in prompt-learning paradigm. Compared to (b), the proposed ViPT has a more concise network structure, benefits from parameter-friendly prompt-tuning, and narrows the gap between the foundation and downstream models.}
	\vspace{-4mm}
	\label{fig:motivation}
\end{figure}

However, as the downstream task of RGB-based tracking, the main issue encountered by multi-modal tracking is the lack of large-scale datasets.
For example, the widely used RGB-based tracking datasets, GOT-10k~\cite{got10k}, TrackingNet~\cite{trackingnet}, and LaSOT~\cite{lasot}, contain 9.3K, 30.1K, and 1.1K sequences, corresponding to 1.4M, 14M, and 2.8M frames for training. 
Whereas the largest training datasets in multi-modal tracking, DepthTrack~\cite{depthtrack}, LasHeR~\cite{lasher}, VisEvent~\cite{visevent}, contain 150, 979, 500 training sequences, corresponding to 0.22M, 0.51M, 0.21M annotated frame pairs, which is at least an order of magnitude less than the former. 
Accounting for the above limitation, multi-modal tracking methods~\cite{depthtrack,dapnet,visevent} usually utilize pre-trained RGB-based trackers and perform fine-tuning on their task-oriented training sets (as shown in Figure~\ref{fig:motivation} (a)$\rightarrow$(b)).
DeT~\cite{depthtrack} adds a depth feature extraction branch to the original ATOM~\cite{atom} or DiMP~\cite{dimp} tracker and fine-tunes on RGB-D training data.
Zhang \etal ~\cite{siamcda} extend SiamRPN++~\cite{siamrpnplusplus} with dual-modal inputs for RGB-T tracking. 
They first construct a unimodal tracking network trained on RGB data, then tune the whole extended multi-modal network with RGB-T image pairs. 
Similarly, Wang \etal ~\cite{visevent} develop dual-modal trackers by extending single-modal ones with various fusion strategies for visible and event flows and perform extra model training on the RGB-E sequences. 
Although effective, the task-oriented full-tuning approach has some drawbacks.
\textit{(i) Full fine-tuning the model is time expensive and inefficient, and 
	the burden of parameters storing is large, which is unfriendly to numerous applications and cumbersome to transfer deploying.}
\textit{(ii) Full fine-tuning is unable to obtain generalized representation due to the limited annotated samples, the inability to utilize the pre-trained knowledge of the foundation model trained on large-scale datasets.}
Thus, a natural question is thrown up: Is there a more effective manner to adapt the RGB-based foundation model to downstream multi-modal tracking?

More recently, in Natural Language Processing (NLP) field, researchers have injected textual prompts into downstream language models to effectively exploit the representational potential of foundation models, this method, is known as prompt-tuning.
After that, a few researchers have tried to freeze the entire upstream model and add only some learnable parameters to the input side to learn valid visual prompts.
Existing researches ~\cite{clip, vpt, doprompt, bahng2022visual} show its great potential and visual prompt learning is expected to be an alternative to full fine-tuning.
Intuitively, there is a large inheritance between multi-modal and single RGB-modal tracking, which should share most of prior knowledge on feature extraction or attention patterns.
In this spirit, we present \textit{ViPT}, a unified visual prompt-tuning paradigm for downstream multi-modal tracking. 
Instead of fully fine-tuning an RGB-based tracker combined with an auxiliary-modal 
branch, ViPT freezes the whole foundation model and only learns a few modal-specific visual prompts, which inherits the RGB-based model parameters trained at scale to the maximum extent (see Figure~\ref{fig:motivation} (c)).
Different from the prompt learning of other single-modal vision tasks, ViPT introduces additional auxiliary-modal inputs into the prompt-tuning process, adapting the foundation model to downstream tasks while simultaneously learning the association between different modalities. 
Specifically, ViPT inserts several simple and lightweight modality-complementary prompter (MCP) blocks into the frozen foundation model to effectively learn the inter-modal complementarities. 
Notably, ViPT is a general framework for various downstream multi-modal tracking tasks, including RGB-D, RGB-T, and RGB-E tracking. 
We summarize the contribution of our work as follows:
\begin{itemize}
	\setlength{\itemsep}{0pt}
	\setlength{\parsep}{0pt}
	\setlength{\parskip}{0pt}
	\item A visual prompt tracking framework is proposed to achieve task-oriented multi-modal tracking. 
	Facilitated by learned prompts, the off-the-shelf foundation model can be effectively adapted from RGB domain to downstream multi-modal tracking tasks. 
	Besides, ViPT is a general method that can be applied to various tasks, \ie, RGB-D, RGB-T, and RGB-E tracking.
	\item A modality-complementary prompter is designed to generate valid visual prompts for the task-oriented multi-modal tracking. 
	The auxiliary-modal inputs are streamlined to a small number of prompts instead of designing an extra network branch.
	\item Extensive experiments show that our method achieves SOTA performance on multiple 
	downstream multi-modal tracking tasks while maintaining parameter-efficient ($<$1\% trainable parameters).
\end{itemize}

\section{Related Work}
\subsection{Multi-Modal tracking}

\begin{figure*}[ht]
	\includegraphics[width=0.99\textwidth]{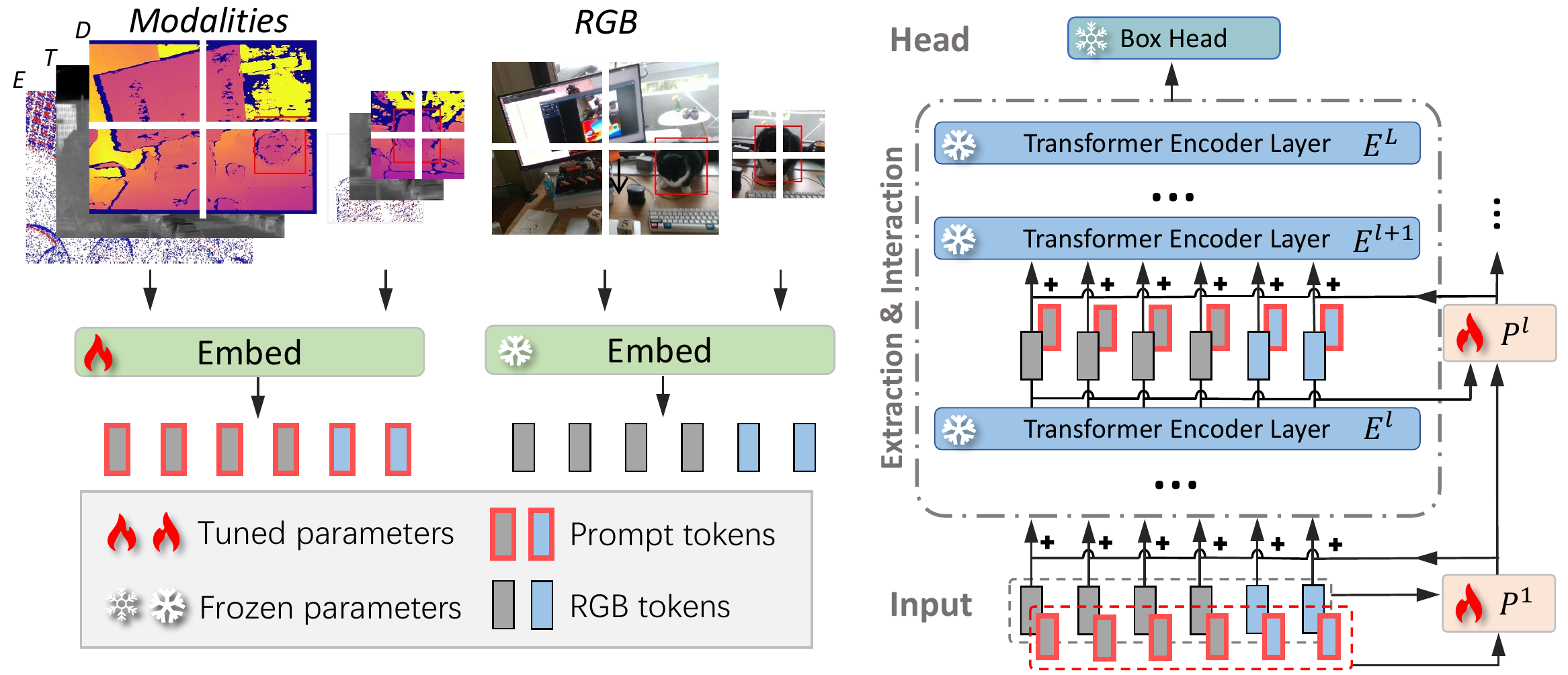}
	\vspace{-1mm}
	\caption{\textbf{Overview architecture of our ViPT.} 
		The RGB- and auxiliary- modal inputs are first fed to the patch embed to generate the corresponding RGB and prompt tokens. 
		$L$-layer stacked vision transformer (ViT) backbone is used for feature extraction and interaction. Modality-complementary prompters $\{P^l\}_{l=1}^{L}$ are inserted into the foundation model to learn the effective visual prompts. 
	}
	\label{fig:overview}
\end{figure*}

Assigned an arbitrary object in the initial frame, general visual object tracking (single-modal inputs) aims to predict the position and scale of the object in all subsequent frames.
The continued emergence of large-scale datasets~\cite{got10k, trackingnet, lasot}, and the surging development of deep neural networks~\cite{resnet,attention_is_all} have further advanced this field.
Numerous trackers \cite {siamesefc, siameserpn, dimp, transt} have been deployed to refresh the accuracy and robustness of tracking.
Even though the existing performance is already impressive, single-modal object tracking in some scenarios (\eg, background clutter, illumination variation, motion blur, \etc.) is still insufficient to meet our demands.
As the cost of binocular imaging sensors becomes increasingly low, multi-modal tracking has attracted more
attention for it can provide complementary information among different modalities, assisting trackers to  handle challenging situations that cannot be addressed solely through RGB inputs.
For example, ~\cite{mueller2017real} assists with additional depth information for hand tracking in occlusion scenes. Similarly, JMMAC~\cite{jmmac} fuses two modalities, RGB and thermal infrared, to obtain reliable appearance and motion cues.
To obtain more robust tracking results, Liu et al. ~\cite{liu2016combined} combines RGB and event flows to predict candidate ROIs in complicated scenarios. 
In addition, numerous works focus on how to perform effective feature fusion on multi-modal information ~\cite{conaire2008thermo, fc2019fusion, fenet, luo2019thermal}.

However, an issue has been overlooked.
A potential risk of overfitting is rising due to the contradiction between the paucity of large-scale multi-model training sets and the huge appetites of the data-driven models.
Therefore, existing methods~\cite{depthtrack,rgbd1k,mfdimp,visevent,dapnet,ca3dms,zhang2021learning} usually design an additional network branch for the inputs of the auxiliary modality, and the final model first loads the pre-trained parameters (from RGB-based model or imagenet~\cite{imagenet}) then conducts fine-tuning on their downstream task-oriented training sets. 
In addition, zhang \etal~\cite{mfdimp} transfer the RGB videos in GOT10k~\cite{got10k} to synthetic TIR videos, and Yan \etal~\cite{depthtrack} generate synthetic RGB-D data from LaSOT~\cite{lasot} and COCO~\cite{coco}.
They attempted to exploit some model-transform algorithms (\eg, ~\cite{miangoleh2021boosting} for monocular depth estimation) to generate auxiliary-modal pseudo-truths from RGB sequences to expand the downstream data, whereas this approach just partly alleviates the data anxiety for multi-modal tracking.
In this work, taking a different path, we design a parameter-efficient prompt-tuning architecture to explore the adaptation from pre-trained RGB-based model to downstream multi-modal tracking tasks.

\subsection{Visual Prompt Learning}
For a long period, fine-tuning has been adopted to leverage large pre-trained models for performing downstream tasks. 
Researchers usually update all the model parameters when training on downstream data. This manner is parameter-inefficient and requires many repetitive task-oriented copies and storage of the entire pre-trained model. 
Recently, as a new paradigm, prompt learning has emerged and drastically boosted the performance of many downstream natural language processing (NLP) tasks~\cite{lester2021power,liu2021pre}. 
Besides, prompt learning also shows its effectiveness in many computer vision 
tasks.
VPT~\cite{vpt} prepends a set of learnable parameters to transformer encoders and remarkably beats full fine-tuning on 20 downstream recognition tasks. 
AdaptFormer~\cite{adaptformer} introduces lightweight modules to ViT and outperforms full fine-tuned models on action recognition benchmarks. 
Convpass~\cite{convpass} is proposed to employ convolutional bypasses for prompt learning on pre-trained ViT. 
ProTrack~\cite{protrack} first introduces the prompting concept into the tracking field.
However, it merely simply fuses source images from two modalities by weighted addition without the tuning process.
The prompts in it are not learnable and thus can gain limited improvements. 
In this work, we exploit the core idea of visual prompt-tuning and design an innovative prompt-learning framework for multi-modal tracking to replace the full fine-tuning paradigm.

\section{Methodology}
\label{sec:method}

In this work, we propose ViPT for effectively and efficiently adapting the pre-trained RGB-based foundation tracking model to downstream multi-modal tasks.
Instead of fully fine-tuning upon a pre-trained foundation model, ViPT only tunes a small number of prompt-learning parameters and achieves promising transfer learning abilities and preferable modal complementarities.
The overall architecture of our ViPT is depicted in Figure~\ref{fig:overview}.

\subsection{Preliminary and Notation}

\textbf{Problem Setting.} 
Given a video with an initial target box $\bm{B}_0$, the goal of the RGB-based tracking is to learn a tracker $F_{RGB}: \{\bm{X}_{RGB}, \bm{B}_0\} \rightarrow \bm{B}$ to predict the box $\bm{B}$ of the target in all subsequent search frames 
${\bm{X}_{RGB}}$.
For multi-modal tracking, it introduces an extra spatial-temporal synchronized input flow, extends the model input to $(\bm{X}_{RGB}, \bm{X}_{A})$, where the subscript $A$ signifies the other auxiliary modalities (\eg, depth, thermal infrared, or event).
Accordingly, the multi-model tracking can be formulated as $F_{MM}: \{\bm{X}_{RGB}, \bm{X}_{A}, \bm{B}_0\} \rightarrow \bm{B}$, where $F_{MM}$ is the multi-modal tracker.

\textbf{Foundation Model.} 
Typically, the tracker $F_{RGB}$ can be decomposed into $\phi\circ f$, where $f: \{\bm{X}_{RGB}, \bm{B}_0\} \rightarrow \bm{\mathcal{H}}_{RGB}$ represents the feature extraction and interaction function, and box head $\phi: \bm{\mathcal{H}}_{RGB} \rightarrow \bm{B}$ estimates the final results.
Hereby, $f$ is a powerful transformer backbone (\ie, ViT~\cite{vit}) in our case. 
Specifically, the input exemplar ${\bm{Z}_{RGB}}$ and search frames $\bm{X}_{RGB}$ are first embedded into patches and flattened to 1D tokens $\bm{\mathcal{H}}_{RGB}^{z}\in\mathbb{R}^{N_z\times D}$ and $\bm{\mathcal{H}}_{RGB}^{x}\in\mathbb{R}^{N_x\times D}$ with positional embedding~\cite{attention_is_all}, where $N_z$ and $N_x$ symbolize the token number of the exemplar and search inputs, and $D$ is the token dimension.
Then the token sequences are concatenated to $\bm{\mathcal{H}}_{RGB}^0=[\bm{\mathcal{H}}_{RGB}^{z}, \bm{\mathcal{H}}_{RGB}^{x}]$.
We feed the token sequences to an $L$-layer standard visual transformer encoder.
Here we denote $\bm{\mathcal{H}}_{RGB}^{l-1}$ as inputs to the $l$-th encoder
layer $E^l$.
The forward propagation process is formulated as:
\begin{align}
	\label{eq:vit}
	&\bm{\mathcal{H}}_{RGB}^l = E^l(\bm{\mathcal{H}}_{RGB}^{l-1}),  &&l=1, 2, \ldots, L \\
	&\bm{B} = \phi(\bm{\mathcal{H}}_{RGB}^L),
\end{align}
where the transformer encoder layer $E^l$ includes Multi-head Self-Attention (MSA), LayerNorm (LN), Feed-Forward Network (FFN), and residual connection, and $\bm{\mathcal{H}}_{RGB}^L$ is the output of the last encoder layer.
The encoder network completes the feature extraction and interaction of exemplar and search patches. 
We refer readers to OSTrack~\cite{ostrack} for more details about our RGB-based foundation model. 

\begin{figure}[tb]
	\centering
	\includegraphics[width=0.47\textwidth]{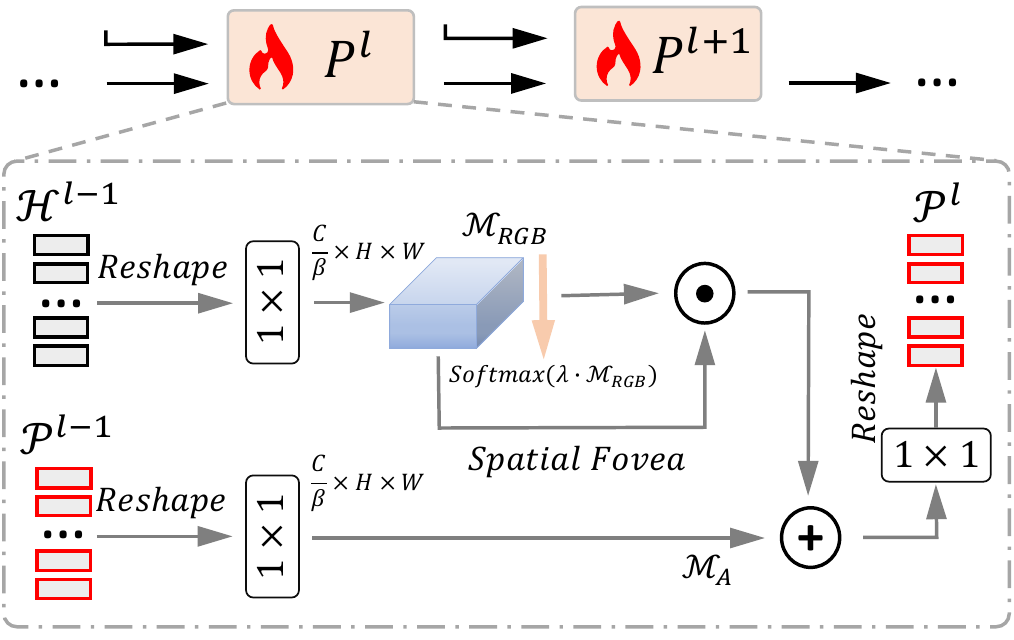}
	\caption{\textbf{Detailed design of the proposed MCP.} 
		Multi-modal input flows are sent to the 1$\times$1 convolutions for distilling to the lower-dimension latent space separately.
		Foundation embeddings first conduct an spatial fovea operation, and then be fused with auxiliary embeddings by element-wise summation. Finally, the multi-modal hybrid embeddings are mapped to the original dimension through a 1$\times$1 convolutional layer.
	}
	\label{fig:prompt}
\end{figure}

\subsection{Multi-Modal Prompt Tracking}
\textbf{Overall Architecture.}
Formally, multi-modal tracking provides an extra auxiliary-modal flow which is temporally synchronized and spatially aligned with the RGB flow.
As shown in Figure~\ref{fig:overview}, ViPT first feeds the two flows ($\bm{X}_{RGB}$ and $\bm{X}_{A}$) to a patch embed layer separately. 
Each input flow is mapped and flattened into $D$-dimensional latent space, we denote them as RGB tokens $\bm{\mathcal{H}}_{RGB}^{0}$ and auxiliary tokens $\bm{\mathcal{H}}_{A}^{0}$.
Then, $\bm{\mathcal{H}}_{RGB}^{0}$ is fed to the foundation model,  $\bm{\mathcal{H}}_{RGB}^{0}$ and $\bm{\mathcal{H}}_{A}^{0}$ are sent into cascade modality-complementary prompter (MCP) to generate modality-specific prompts. 
The learned prompts are added to the original RGB flow in a form of residuals:
\begin{align}
	\label{eq:prompt_0}
	&\bm{\mathcal{H}}^{l} =\bm{\mathcal{H}}^{l}_{RGB} + \bm{\mathcal{P}}^{l+1}, &&l=0, 1, \ldots, L-1
\end{align}
where $\bm{\mathcal{H}}^{l}$ is the prompted tokens that will be injected into the next layer of the foundation model,
and $\bm{\mathcal{P}}^{l+1}$ symbolizes prompt token sequences from the $(l+1)$-th MCP block.
Hereby, we place stage-wise MCP blocks to take full advantage of semantic understanding of diverse levels and diverse modalities.
Directly adding prompts to the intermediate features of the foundation model also enables our ViPT to be quickly and easily applied to existing pre-trained foundation trackers.
It is worth noting that, different from prompt-tuning approaches (\eg, SPM~\cite{spm}) which contain trainable prompt-learning network and prediction head,
in our ViPT, all the RGB-modal relevant network parameters are frozen, including the patch embedding, feature extraction and interaction, and prediction head. 

\textbf{Modality-Complementary Prompter.}
Recently, a few works begin to explore introducing some trainable parameters to the frozen pre-trained model to learn effective visual prompts. 
By fine-tuning only a small number of parameters for prompt learning, they achieve impressive performance on a wide range of vision tasks.
A more challenging task, however, is not only to close the gap between upstream and downstream tasks but also to make appropriate and effective harness of inter-modal information.
As shown in Figure~\ref{fig:overview}, the proposed MCP blocks are inserted into multiple stages of the foundation network.
Formally, providing token sequences $\{\bm{\mathcal{H}}_{RGB}^0,\bm{\mathcal{H}}_{A}^0\}$ of two modalities from a downstream task, and a frozen foundation encoder $f_{*}$ containing $L$ transformer layers $\{E_{*}^{l}\}_{l=1}^{L}$, 
the designed MCP is to learn the prompts with these two input flows, the process can be written as:
\begin{align}
	\label{eq:prompt}
	&\bm{\mathcal{P}}^l = P^{l}(\bm{\mathcal{H}}^{l-1}, \bm{\mathcal{P}}^{l-1}), &l=1, 2, \ldots, L 
\end{align}
where $P^{l}$ denotes the $l$-th MCP block.
Specifically, $\bm{\mathcal{H}}^{0}=\bm{\mathcal{H}}^{0}_{RGB}$ and $\bm{\mathcal{P}}^{0}=\bm{\mathcal{H}}^{0}_{A}$. 
In this way, MCP sufficiently extracts the feature representations of the downstream task at different semantic levels, while learning the complementarity between the two modalities and generating more robust prompts. 
The blending representation can contribute to a complementarity between the intermediate foundation feature and the learned prompts, and a balance between the foundation and auxiliary modalities.

The detailed design of MCP is shown in Figure~\ref{fig:prompt}, our MCP has two input branches for adhibiting the token sequences of the foundation flows $\bm{\mathcal{H}}^{l-1}$ and auxiliary flows $\bm{\mathcal{P}}^{l-1}$, respectively.
MCP comprises three main steps:
(i) projecting to lower-dimensional latent embeddings; 
(ii) generation of the multi-modal inner complementary representations; 
and (iii) projecting to original dimension and generating the learned multi-modal visual prompts.
Concretely, considering that single-modal flow has certain redundant features, and the prompt block should have as few parameters as possible for practical purposes, we project each flow into $[\frac{C}{\beta}\times H\times W]$ dimension by:
\begin{align}
	\label{eq:project}
	\bm{\mathcal{M}}_{RGB} = g_{1}(\bm{\mathcal{H}}^{l-1}), \quad \bm{\mathcal{M}}_{A} = g_{2}(\bm{\mathcal{P}}^{l-1}), 
\end{align}
in our case, the input channel $C$ is 768, and the reducing factor $\beta$ is set to $\frac{768}{8}$ in all the prompt blocks.
Hereby, the projection functions $g_1(\cdot)$ and $g_2(\cdot)$ are  simple 1$\times$1 convolutional layers. 
The foundation embeddings $\bm{\mathcal{M}}_{RGB}$ then perform the spatial fovea operation, 
which first applies a $\lambda$-smoothed spatial $softmax$ across all the spatial dimensions, and produces the enhanced embeddings $\bm{\mathcal{M}}_{RGB}^{e}$ by applying the channel-wise spatial attention-like mask $\bm{\mathcal{M}}_{fovea}$ over $\bm{\mathcal{M}}_{RGB}$,
\begin{align}
	\label{eq:fovea}
	&\bm{\mathcal{M}}_{RGB}^{e} = \bm{\mathcal{M}}_{RGB} \odot \bm{\mathcal{M}}_{fovea}, \\ 
	&\bm{\mathcal{M}}_{fovea} = \{\frac{e^{\bm{\mathcal{M}}_{RGB}^{[:,i,j]}}}{\Sigma e^{\bm{\mathcal{M}}_{RGB}^{[:,i,j]}}}\lambda\},
\end{align} 
where $i=1,2,...,H$ and $j=1,2,...,W$, and $\lambda$ is a learnable weighted parameter per block.
Then, we obtain mixed-modal embeddings by additive binding,
and the learned prompts can be gained by,
\begin{align}
	\label{eq:mix_emb}
	\bm{\mathcal{P}}^{l} &= g_3(\bm{\mathcal{M}}_{RGB}^{e} + \bm{\mathcal{M}}_{A}),
\end{align} 
where $g_3(\cdot)$ is the same as $g_1(\cdot)$ and $g_2(\cdot)$.

\textbf{Optimization.}
The multi-modal tracking model $F_{MM}$ is parameter-initialized by the foundation model $F_{RGB}$.
During prompt-tuning, data flows propagate throughout the entire model but we only update the gradient values of a few parameters 
$\theta = \{ \tau^{A}, \{P^{l}\}_{l=1}^{L} \}$ for visual prompt learning, where $\tau^{A}$ denotes the patch embedding functions of auxiliary inputs. 
Therefore, the default version of our ViPT contains only 0.84M trainable parameters. 
Despite this, we find that ViPT defeats the full fine-tuning paradigm with two branches for various modalities.
Thus, the optimization process can be formulated as
\begin{align}
	\label{eq:optimization}
	\theta_{tuned}&=\arg \underset{\theta}{\min } \frac{1}{\left|{\mathcal{D}}\right|} \sum \mathcal{L} \left(\phi (\bm{\mathcal{H}}^{L}), \bm{B}_{gt}\right),
\end{align}
where $\mathcal{D}$ symbolizes the downstream multi-modal data.
By tuning only a few parameters for prompt learning, the model can achieve convergence within a brief period, and effectively leverage the abundant knowledge in the pre-trained foundation model, thereby narrowing the gap between the models.
The overall loss function of ViPT is the same as the foundation model without extra adjusting, 
\begin{align}
	\label{eq:loss}
	\mathcal{L}=L_{cls}+\lambda_{iou}L_{iou}+\lambda_{L_1}L_1.
\end{align}
where $L_{cls}$ is the weighted focal loss for classification, $L_1$ and generalized IoU loss $L_{iou}$ are employed for bounding box regression, $\lambda_{iou}$ and $\lambda_{L_1}$ are the regularization parameters, and all the corresponding settings are the same as ~\cite{ostrack}.

\subsection{Discussion}

\begin{table*}[!ht]
	\small
	\centering
	\renewcommand\arraystretch{1.15} 
	\setlength{\tabcolsep}{1.5mm}{
		\resizebox{\linewidth}{!}{
			\begin{tabular}{c|cccccccccccccc}
				\hline
				\small
				&\tabincell{c}{CA3DMS\\~\cite{ca3dms}}&\tabincell{c}{SiamM\_Ds\\~\cite{vot19}}&\tabincell{c}{Siam\_LTD\\~\cite{vot20}}&\tabincell{c}{LTDSEd\\~\cite{vot19}}&\tabincell{c}{DAL\\~\cite{dal}}&\tabincell{c}{CLGS\_D\\~\cite{vot20}}&\tabincell{c}{LTMU\_B\\~\cite{vot20}}&\tabincell{c}{ATCAIS\\~\cite{vot20}}&\tabincell{c}{DDiMP\\~\cite{vot20}}&\tabincell{c}{DeT\\~\cite{depthtrack}}&\tabincell{c}{OSTrack\\~\cite{ostrack}}&\tabincell{c}{SPT\\~\cite{rgbd1k}}&\tabincell{c}{ProTrack\\~\cite{protrack}}&\tabincell{c}{\textbf{ViPT}\\\textbf{(ours)}}\\
				\hline
				F-score($\uparrow$)&0.223&0.336&0.376&0.405&0.429&0.453&0.460&0.476&0.485&0.532&0.529&\textbf{\textcolor[rgb]{0,0,1}{0.538}}&\textbf{\textcolor[rgb]{0,1,0}{0.578}}&\textbf{\textcolor[rgb]{1,0,0}{0.594}} \\
				Re($\uparrow$)&0.228&0.264&0.342&0.382&0.369&0.369&0.417&0.455&0.469&0.506&0.522&\textbf{\textcolor[rgb]{0,0,1}{0.549}}&\textbf{\textcolor[rgb]{0,1,0}{0.573}}&\textbf{\textcolor[rgb]{1,0,0}{0.596}} \\
				Pr($\uparrow$)&0.218&0.463&0.418&0.430&0.512&0.584&0.512&0.500&0.503&\textbf{\textcolor[rgb]{0,0,1}{0.560}}&0.536&0.527&\textbf{\textcolor[rgb]{0,1,0}{0.583}}&\textbf{\textcolor[rgb]{1,0,0}{0.592}} \\
				\hline
			\end{tabular}
	}}\\
	\vspace{-2mm}
	\caption{
		\small
		Overall performance on DepthTrack test set~\cite{depthtrack}.}
	\label{tab-depthtrack}
	\vspace{-1mm}
\end{table*}

\begin{table*}[htbp]
	\small
	\centering
	\renewcommand\arraystretch{1.15} 
	\setlength{\tabcolsep}{1.5mm}{ 
		\resizebox{\linewidth}{!}{
			\begin{tabular}{c|ccccccccccccc}
				\hline
				\small
				&\tabincell{c}{ATOM\\~\cite{atom}}&\tabincell{c}{DiMP\\~\cite{dimp}}&\tabincell{c}{ATCAIS\\~\cite{vot20}}&\tabincell{c}{DRefine\\~\cite{vot21}}&\tabincell{c}{keep\_track\\~\cite{vot22}}&\makebox[0.075\textwidth][c]{\tabincell{c}{STARK\_RGBD\\~\cite{vot21}}}&\tabincell{c}{DMTracker\\~\cite{vot22}}&\tabincell{c}{DeT\\~\cite{depthtrack}}&\tabincell{c}{OSTrack\\~\cite{ostrack}}&\tabincell{c}{SBT\_RGBD\\~\cite{vot22}}&\tabincell{c}{SPT\\~\cite{rgbd1k}}&\tabincell{c}{ProTrack\\~\cite{protrack}}&\tabincell{c}{\textbf{ViPT}\\\textbf{(ours)}}\\
				\hline
				EAO($\uparrow$)&0.505&0.543&0.559&0.592&0.606&0.647&0.658&0.657&\textbf{\textcolor{blue}{0.676}}&\textbf{\textcolor{green}{0.708}}&0.651&0.651&\textbf{\textcolor[rgb]{1,0,0}{0.721}} \\
				Accuracy($\uparrow$)&0.698&0.703&0.761&0.775&0.753&\textbf{\textcolor{blue}{0.803}}&0.758&0.760&\textbf{\textcolor{blue}{0.803}}&\textbf{\textcolor{green}{0.809}}&0.798&0.801&\textbf{\textcolor[rgb]{1,0,0}{0.815}} \\
				Robustness($\uparrow$)&0.688&0.731&0.739&0.760&0.797&0.798&\textbf{\textcolor{blue}{0.851}}&0.845&0.833&\textbf{\textcolor{green}{0.864}}&\textbf{\textcolor{blue}{0.851}}&0.802&\textbf{\textcolor[rgb]{1,0,0}{0.871}} \\
				\hline
			\end{tabular}
	}}
	\vspace{-2mm}
	\caption{
		\small
		Overall performance on VOT-RGBD2022~\cite{vot22}.}
	\label{tab-vot22rgbd}
	\vspace{-1mm}
\end{table*}

\textbf{The potential of prompt-tuning tracking.}
We believe that the prompt-tuning paradigm has great potential for multi-modal tracking, mainly in terms of \textit{(i) Prompt-tuning has better adaptability than full fine-tuning, especially for downstream multi-modal tracking where large-scale data is scarce.}
Full fine-tuning on the downstream dataset corrupts the quality of pre-trained parameters,
so the trackers are more likely to over-fit or reach a sub-optimal state. 
Especially in the task of RGB+auxiliary modality tracking, the original foundation knowledge can be used directly for the extraction of RGB-modal features.
Prompt-tuning fixes the parameters of the foundation model, retaining the knowledge of the pre-trained model, and the prompt module allows for flexible adaptation to task-oriented data.
\textit{(ii) Prompt-tuning allows for a closer association between RGB and RGB+auxiliary modality tracking, as well as learning about the modal complementarities.}
RGB and auxiliary modalities have different data distributions, so providing an additional feature extraction network for auxiliary-modal inputs may reduce the inter-modal connectivity.
\textit{(iii) The remarkable advantages of practical application.}
Prompt-tuning tracking has significantly fewer trainable parameters than full fine-tuning, often by an order of magnitude or more.
In our work, ViPT contains only 0.84M trainable parameters, less than 1\% of the whole model parameters (93.36M).
The difference in trainable parameters is further reflected in the more practical aspects of application.
Compared with full fine-tuning, prompt-tuning has a shorter training period,
at the same time, this paradigm can be deployed to various downstream tracking scenarios without having to store a huge number of foundation parameters repeatedly.
These advantages further enables the foundation tracker to be quickly transferred to diverse downstream tracking tasks.

\textbf{Differences with other prompt-learning methods.}
The proposed ViPT focuses on effectively introducing prompt-learning mechanisms to multi-modal tracking.
In contrast to the existing prompt-learning methods in CV or NLP domains which focus on adaptation between upstream to downstream tasks, ViPT also explores adaptation between single-modal and multi-modal tasks.
A similar work to ours is ProTrack~\cite{protrack}. 
Unfortunately, ProTrack merely introduces the concept of prompt to multi-modal tracking.
Specifically, it conducts a linear summing of multi-modal inputs and uses only RGB-based model parameters without tuning on the downstream data.
In contrast, ViPT fully explores the associations between different modalities at different semantic levels, learns the complementarities between modalities in the form of prompt-learning, therefore achieving superior performance.

\section{Experiments}

\subsection{Downstream tasks}

ViPT achieves the unification of multiple downstream multi-modal tracking tasks.
In this work, three challenging tasks are chosen to validate the effectiveness and generalization of the proposed method.
(i) For RGB-D tracking, we evaluate our tracker on Depthtrack~\cite{depthtrack} and VOT22-RGBD~\cite{vot22}.
(ii) For RGB-T tracking,
we provide the comparison results on RGBT234~\cite{rgbt234} and LasHeR~\cite{lasher}. 
(iii) For RGB-E tracking, we report the experimental results on the largest VisEvent~\cite{visevent}.
We perform prompt-tuning on these downstream tasks without specific modulation, and all the experimental settings are kept to the same.

\subsection{Experimental Settings}
When fine-tuning our ViPT on downstream tasks, we choose train sets of Depthtrack~\cite{depthtrack} for RGB-D tracking, LasHeR~\cite{lasher} for RGB-T tracking, and VisEvent~\cite{visevent} for RGB-E tracking.
Some recent works (\eg, DeT~\cite{depthtrack} and mfDiMP~\cite{mfdimp}) employ other algorithms to generate more sample pairs for 
downstream tasks, thus alleviating the problem of insufficient data in downstream tasks.
While in our work, to demonstrate the effectiveness of prompt-tuning, we only use the most basic multi-modal training set mentioned above.
ViPT is trained on 2 NVIDA Tesla A100 GPUs with a global batch size of 64. 
The model fine-tuning takes 60 epochs that each epoch contains $6\times10^{4}$ sample pairs.
The AdamW optimizer~\cite{adamw} with a weight decay of $10^{-4}$ is adopted.
The initial learning rate is set to $4\times10^{-5}$ and decreased by the factor of 10 after 48 epochs.
The fixed parameters are initialized with the pre-trained foundation model~\cite{ostrack} in RGB-based tracking, and the trainable prompt learning parameters are initialized 
with xavier uniform initialization scheme~\cite{xavier}.

\subsection{Main Properties and Analysis}

\begin{figure}[bp]
	\centering
	\vspace{-2.5mm}
	\includegraphics[width=1.0\linewidth]{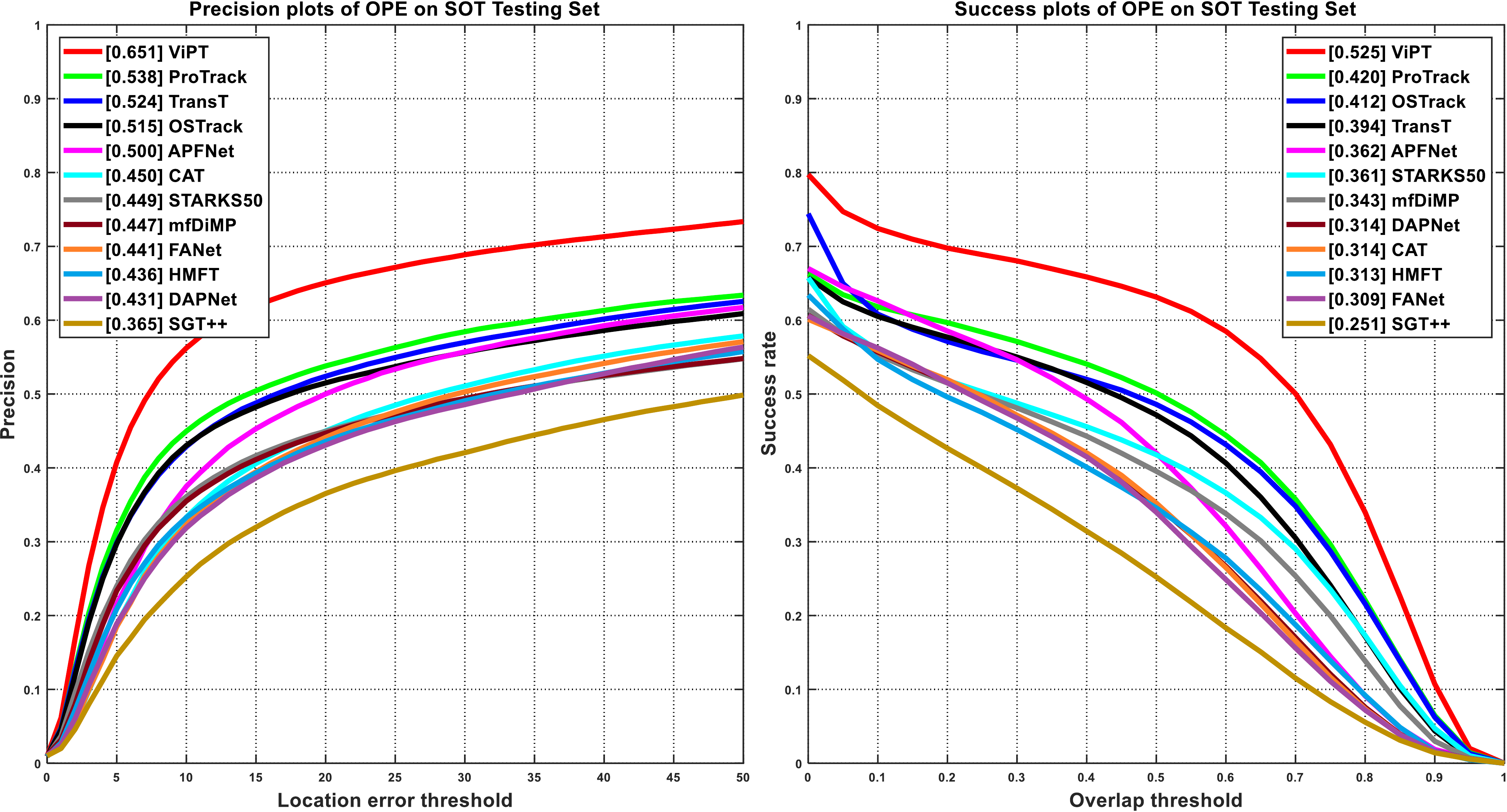}
	\vspace{-6mm}
	\caption{Overall performance on LasHeR test set~\cite{lasher}.}
	\label{fig:lasher}
\end{figure}

\begin{table*}[h]
	\small
	\centering
	\renewcommand\arraystretch{1.15} 
	\resizebox{\linewidth}{!}{
		\begin{tabular}{c|cccccccccccc}
			\hline
			\small
			&\tabincell{c}{mfDiMP\\~\cite{mfdimp}} &\tabincell{c}{SGT\\~\cite{sgt}} &\tabincell{c}{DAFNet\\~\cite{dafnet}} &\tabincell{c}{OSTrack\\~\cite{ostrack}}&\tabincell{c}{FANet\\~\cite{fanet}}  &\tabincell{c}{MaCNet\\~\cite{macnet}} &\tabincell{c}{CAT\\~\cite{cat}} &\tabincell{c}{JMMAC\\~\cite{jmmac}} &\tabincell{c}{CMPP\\~\cite{cmpp}} &\tabincell{c}{APFNet\\~\cite{apfnet}} &\tabincell{c}{ProTrack\\~\cite{protrack}} &\tabincell{c}{\textbf{ViPT}\\\textbf{(ours)}}\\ 
			\hline
			MPR($\uparrow$) &0.646 &0.720 &0.796 &0.729 &0.787 &0.790 &0.804 &0.790 &\textcolor{blue}{\textbf{0.823}} &\textcolor{green}{\textbf{0.827}} &0.795 &\textcolor{red}{\textbf{0.835}} \\
			MSR($\uparrow$) &0.428 &0.472 &0.544 &0.549 &0.553 &0.554 &0.561 &0.573 &0.575 &\textcolor{blue}{\textbf{0.579}} &\textcolor{green}{\textbf{0.599}} &\textcolor{red}{\textbf{0.617}} \\
			\hline
		\end{tabular}
	}
	\vspace{-2mm}
	\caption{
		\small
		Overall performance on RGBT234 dataset~\cite{rgbt234}.}
	\label{tab-rgbt234}
	\vspace{-3mm}
\end{table*}

\textbf{DepthTrack.}                                                       
DepthTrack~\cite{depthtrack} is a large-scale long-term RGB-D tracking benchmark, which contains 150 training and 50 testing videos with 15 per-frame attributes. 
It uses precision (Pr) and recall (Re) to measure the accuracy and robustness of target localization.
F-score, calculated by $F=\frac{2RePr}{Re+Pr}$, is the primary measure. Though our ViPT is a short-term algorithm, Table~\ref{tab-depthtrack} shows ViPT surpasses all previous SOTA trackers and obtains the highest F-score of 59.4\%, gaining a significant improvement of 6.5\% compared with our foundation model~\cite{ostrack}.

\textbf{VOT-RGBD2022.} VOT-RGBD2022~\cite{vot22} is an up-to-date benchmark that contains 127 short-term RGB-D sequences for exploiting the role of depth in RGB-D tracking. 
It applies an anchor-based short-term evaluation protocol introduced in ~\cite{vot20} that needs trackers to multi-start from different initialization points. 
The expected average overlap (EAO) is the overall performance measure. 
As reported in Table~\ref{tab-vot22rgbd}, our ViPT is superior to previous competitive methods, getting an EAO of 0.721. 
ViPT outperforms the foundation model by 4.5\% EAO while another prompt-based tracker ProTrack~\cite{protrack}, only has an improvement of 0.4\% relative to their baseline STARK\_RGBD~\cite{stark}.

\textbf{RGBT234.} RGBT234~\cite{rgbt234} is a large-scale RGB-T tracking benchmark, which contains 234 videos with visible and thermal infrared pairs. 
MSR and MPR are adopted for performance evaluation. 
Here we compare our proposed ViPT with recent RGB-T trackers. 
As shown in Table~\ref{tab-rgbt234}, our ViPT reaches the top MSR and MPR of 61.7\% and 83.5\%, respectively, which beats the well-designed RGB-T trackers and outperforms ProTrack~\cite{protrack} by 1.8\% on MSR.

\textbf{LasHeR.} LasHeR~\cite{lasher} is a large-scale high-diversity benchmark for short-term RGB-T tracking. 
We evaluate the trackers on 245 testing video sequences in terms of precision plot and success plot. 
The results are reported in Figure~\ref{fig:lasher}.
Surprisingly, ViPT surpasses previous SOTA algorithms by a large margin, which exceeds the second place by 10.5\% and 11.3\% on success and precision, respectively. 
It can be inferred that our prompt-tuning paradigm effectively utilizes the thermal information and makes ViPT adapt well to complex tracking scenarios.

\begin{figure}[hp]
	\centering
	\includegraphics[width=1.0\linewidth,height=0.53\linewidth]{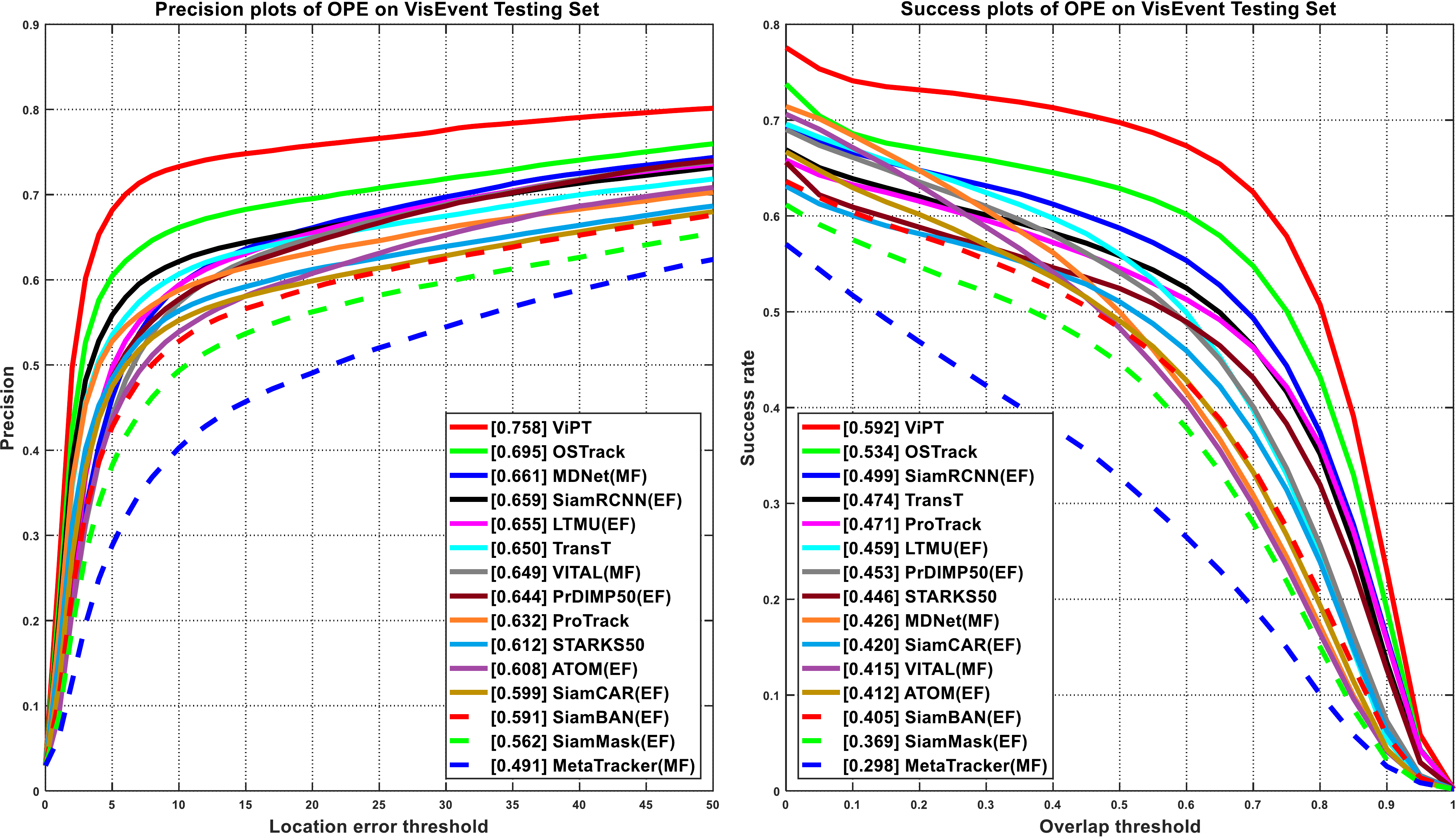}
	\vspace{-5mm}
	\caption{Overall performance on VisEvent test set~\cite{visevent}.}
	\label{fig:visevent}
	\vspace{-1mm}
\end{figure}

\textbf{VisEvent.} VisEvent~\cite{visevent} is the largest visible-event benchmark dataset collected from real-world for single object tracking at present and we perform comparisons on its 320 testing videos. 
Note we just use event images transformed from raw event data rather than event flow. 
Figure~\ref{fig:visevent} shows ViPT achieves a gain of 5.8\% on success and 6.3\% on precision over the runner-up OSTrack~\cite{ostrack} and outperforms other visible-event SOTA trackers. 

\subsection{Exploration Studies}
We further explore the characteristics of our ViPT on multiple downstream tasks of multiple representative benchmarks.
We present the experimental results on Depthtrack~\cite{depthtrack}, LasHeR~\cite{lasher}, and VisEvent~\cite{visevent}.

\begin{figure}[t]
	\centering
	\vspace{0.mm}
	\includegraphics[width=0.4825\textwidth]{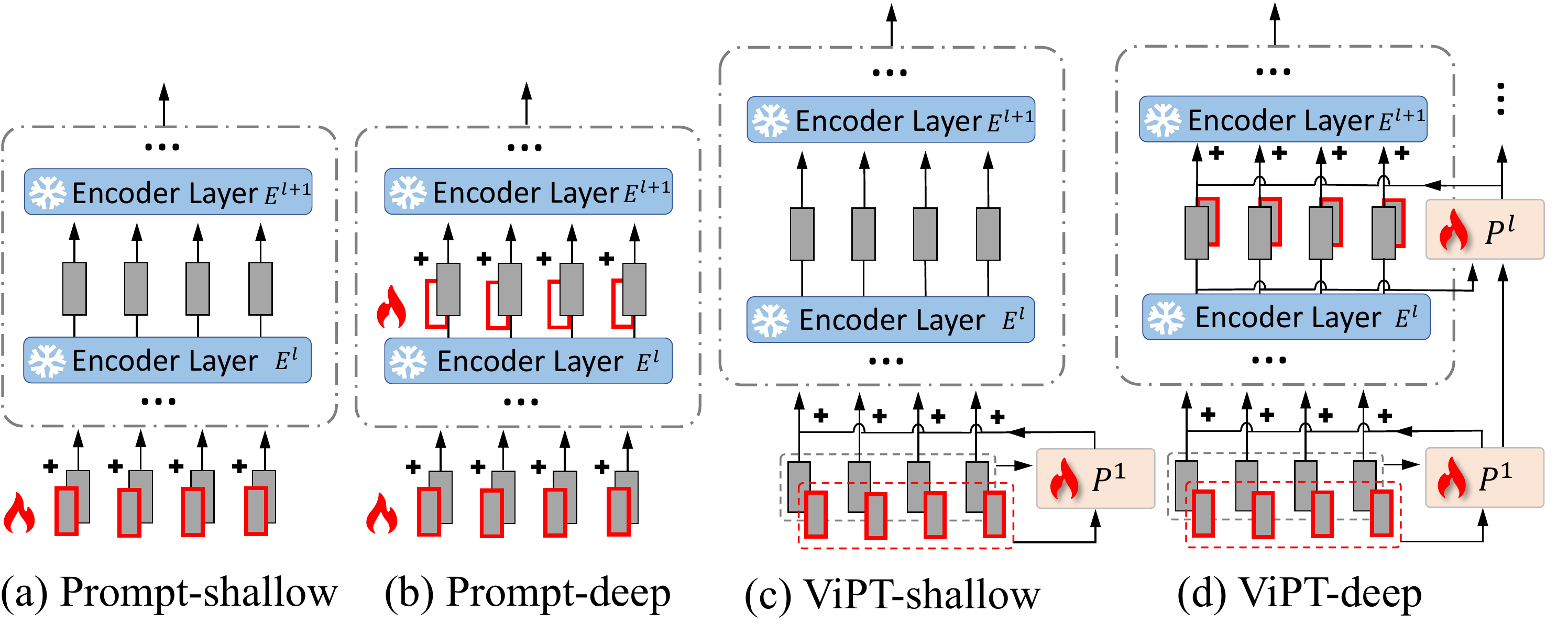}
	\vspace{-6mm}
	\caption{Variants of vanilla prompt-structure and ViPT.}
	\vspace{-4mm}
	\label{fig:variants}
\end{figure}

\textbf{Power of Multiple Modalities.}
To investigate the benefits gained from multi-modal flows, we evaluate the performance with single-modal inputs (RGB-based foundation tracker) and multi-modal inputs. 
For multi-modal cases, we further compare the effectiveness between FFT and our ViPT. 
Specifically, FFT is the full fine-tuning tracker which is extended from the foundation model by adding an auxiliary-modal branch, and the branch design and modalities interaction are inspired by ~\cite{depthtrack}. 
As shown in Table~\ref{tab:power_of_mm}, after combining the auxiliary-modal inputs, FFT shows remarkable improvement over the foundation model, especially on LasHeR (10.5\%$\uparrow$). 
It is encouraging to note that ViPT still beats FFT on performance while having two orders of magnitude fewer trainable parameters.

\begin{table}[ht]
	\renewcommand\arraystretch{1.25}
	\centering
	\fontsize{8}{10}\selectfont  
	\setlength{\tabcolsep}{0.6mm}{
		\resizebox{\linewidth}{!}{%
			\begin{tabular}{c|c|ccc|cc|cc}
				\hline
				\multirow{2}{*}{Method} &
				\multirow{2}{*}{Params$^{\dag}$} &
				\multicolumn{3}{c|}{Depthtrack~\cite{depthtrack}} &
				\multicolumn{2}{c|}{LasHeR~\cite{lasher}} &
				\multicolumn{2}{c}{VisEvent~\cite{visevent}}  \\
				\cline{3-9}
				&& F-score($\uparrow$) & Re($\uparrow$) & Pr($\uparrow$) & SR($\uparrow$) & PR($\uparrow$) & SR($\uparrow$) & PR($\uparrow$)   \\
				\hline
				{Foudation} &- & 52.9 & 52.2 & 53.6 & 41.2 & 51.5 & 53.4 & 69.5  \\
				
				{FFT} &178.6M (100\%)&55.6 (2.7$\uparrow$) &55.4&55.7&51.7 (10.5$\uparrow$) &64.8&58.7 (5.3$\uparrow$) &75.2  \\
				
				{ViPT} &0.84M (0.9\%)& \textbf{59.4}(6.5$\uparrow$)  & \textbf{59.6} & \textbf{59.2} & \textbf{52.5}(11.3$\uparrow$) & \textbf{65.1} & \textbf{{59.2}} (5.8$\uparrow$) &\textbf{75.8}  \\
				
				\hline
				Prompt-shaw &0.59M (0.6\%)&52.0 &51.5 &52.5 &47.2 &58.7 &53.7 &70.2  \\
				
				Prompt-deep &3.29M (3.4\%)&54.9 &54.7 &55.1 &50.7 &63.1 &54.8 &71.4  \\
				
				ViPT-shaw &0.61M (0.7\%)& 56.4 & 56.7& 56.2 &47.9  &59.6  &57.8  &74.9  \\
				
				\hline
	\end{tabular}}}
	\vspace{-2mm}
	\caption{Ablation studies on multiple downstream benchmarks.
		Params$^\dag$ denotes the number of trainable parameters, and its proportion of the overall model parameters is noted in parentheses.
		Results are reported in percentage (\%).}
	\vspace{-0mm}
	\label{tab:power_of_mm}
\end{table}

\textbf{Variants Comparison.}
We further explore different variant versions of our ViPT.
As shown in Figure~\ref{fig:variants} (c) \& (d) , we provide two variants of ViPT, ViPT-shallow (ViPT-shaw in Table~\ref{tab:power_of_mm}) and ViPT-deep (ViPT in Table~\ref{tab:power_of_mm}).
Moreover, to verify the validity of our MCP module, we introduce prompt tokens into the foundation model with VPT-style~\cite{vpt} (Figure~\ref{fig:variants} (a) \& (b)).
Note that for receiving auxiliary-modal inputs,
we replace the input prompts with the embeddings of auxiliary modality.
And we choose to introduce the prompt parameters as a summation instead of the concatenation to align RGB and other auxiliary modalities.
The trainable parameters and performance comparison are reported in Table~\ref{tab:power_of_mm}. 
We can see that ViPT achieves favorable precision-efficiency trade-offs.

\begin{figure}[t]
	\centering
	\includegraphics[width=0.99\linewidth]{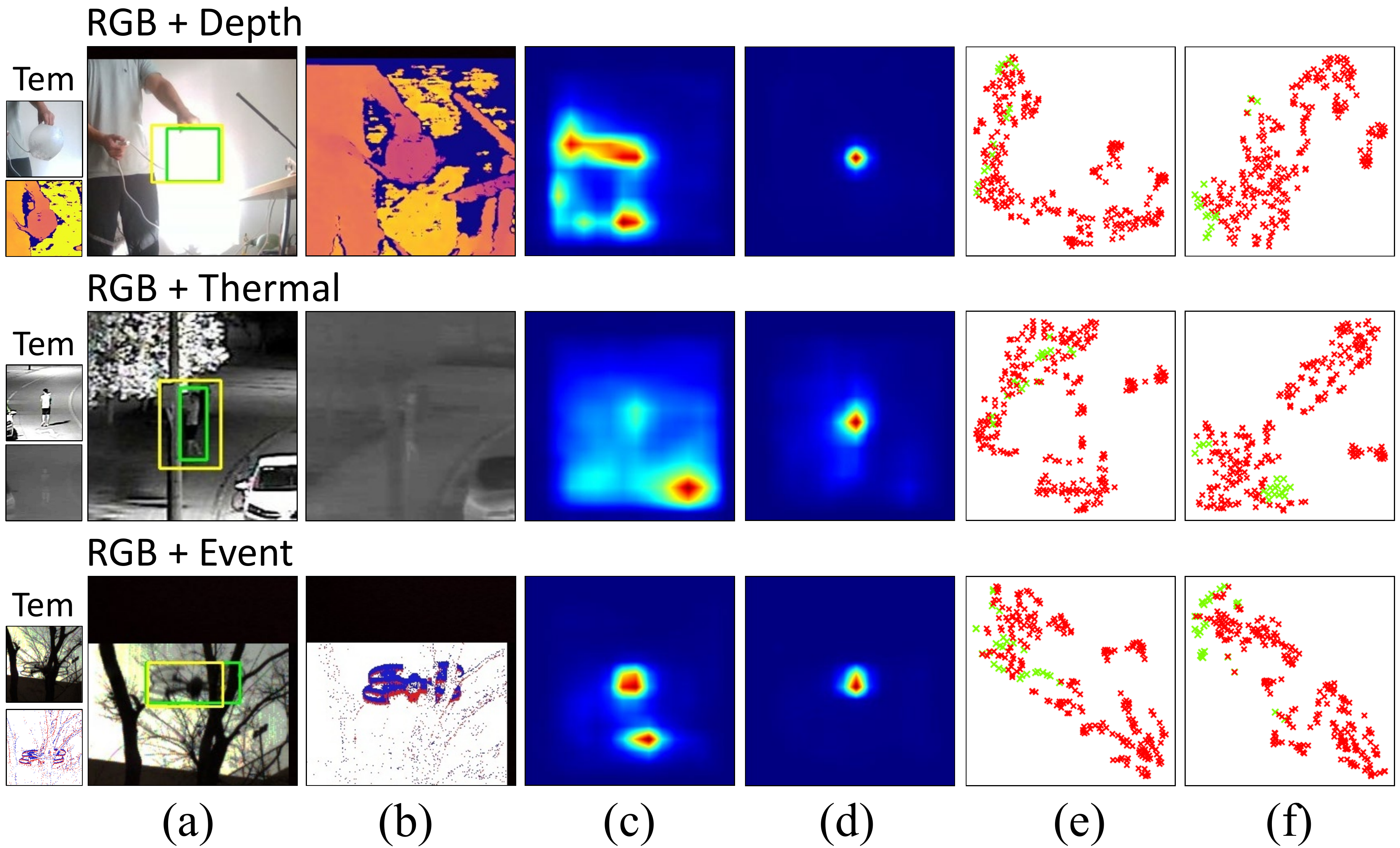}
	\vspace{-2.5mm}
	\caption{Visualization of response and t-SNE~\cite{tsne} maps. (a) RGB flows. The \textcolor{yellow}{yellow} and \textcolor{green}{green} bounding boxes denote Foundation tracker and ViPT, respectively. (b) Auxiliary-modal flows. (c) Score maps of foundation model. (d) Score maps of ViPT. (e) t-SNE maps of the foundation model. (f) t-SNE maps of ViPT.}
	\vspace{-2mm}
	\label{fig:tsne}
\end{figure}

\begin{figure}[!ht]
	\centering
	\begin{subfigure}{0.158\textwidth}
		\centering   
		\includegraphics[width=\linewidth]{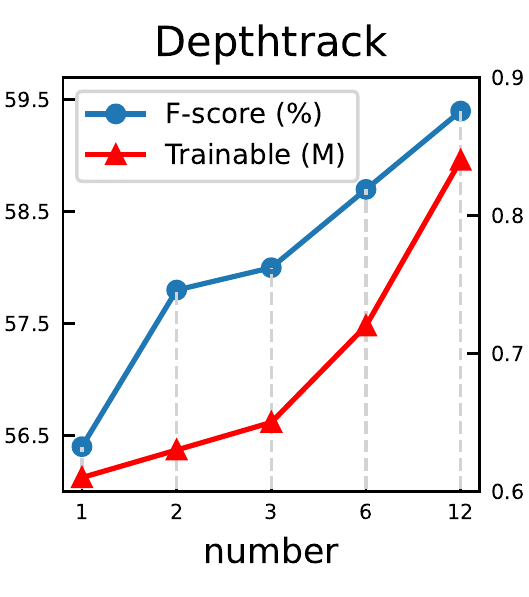}
	\end{subfigure}  
	\hspace{-1.5mm}
	\begin{subfigure}{0.152\textwidth}
		\centering   
		\includegraphics[width=\linewidth]{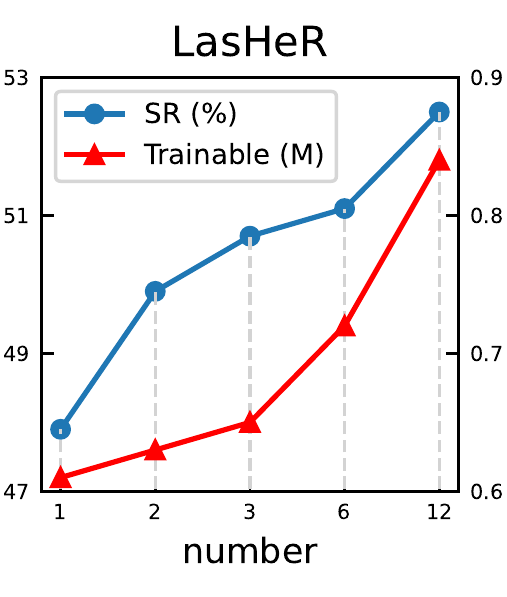}
	\end{subfigure}
	\hspace{-1.5mm}
	\begin{subfigure}{0.158\textwidth}
		\centering   
		\includegraphics[width=\linewidth]{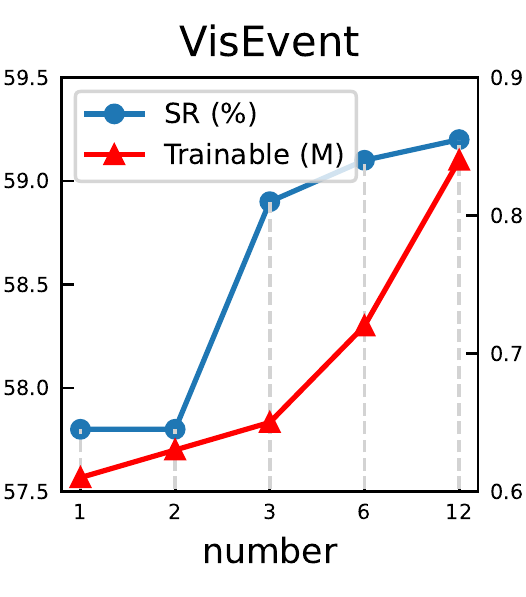}
	\end{subfigure}
	\vspace{-3.5mm}
	\caption{Influence of the number of MCP blocks.}
	\vspace{-2mm}
	\label{fig:prompt_block}
\end{figure}

\textbf{Number Analysis of Prompt Block.}
ViPT achieves multi-modal tracking by inserting MCP blocks into different semantic layers of the foundation model. 
Here we experimentally investigate the effect of providing different numbers of prompters to the foundation model.
Specifically, we set different placement intervals for the inserted blocks, and we insert MCP per 1, 2, 4, 6, and 12 layers of the foundation encoder.
Note that the first and the last case are the variants of our ViPT-deep and ViPT-shallow.
As shown in Figure~\ref{fig:prompt_block}, we observed that as the number of inserted MCP blocks increases, there is a corresponding increase in performance on all benchmarks. 

\textbf{Training Volume and Tunable Parameters.}
We note that DeT~\cite{depthtrack} and mfDiMP~\cite{mfdimp} expand the training data by generating more RGB-D and RGB-T image pairs.
Hence, we further expand the training data to see its impact. In the implementation, we use the same data generation method as them.
Moreover, we unfreeze the overall parameters of ViPT to see the impact of number of trainable parameters.
The results are reported in Table~\ref{tab:expand_data_parameters}.
We can see that only expending the train set or unfreezing the rest parameters can not bring better results.
The reason may be that a small number of trainable parameters is more conducive for model learning in downstream tasks with limited train data.
Conducting them together may gain some benefits (\eg, on LasHeR), but it will be more costly compared with ViPT.

\begin{table}[ht]
	\centering
	\fontsize{8}{10}\selectfont  
	\setlength{\tabcolsep}{1.6mm}{
		\resizebox{\linewidth}{!}{%
			\begin{tabular}{c|cc|ccc|cc}
				\hline
				\multirow{2}{*}{Model} &
				\multirow{2}{*}{ET} &
				\multirow{2}{*}{UFP} &
				\multicolumn{3}{c|}{Depthtrack~\cite{depthtrack}} &
				\multicolumn{2}{c}{LasHeR~\cite{lasher}}  \\
				\cline{4-8}
				&&& F-score($\uparrow$) & Re($\uparrow$) & Pr($\uparrow$) & SR($\uparrow$) & PR($\uparrow$) \\
				\hline
				ViPT && & 59.4 & 59.6 & 59.2 & 52.5 & 65.1  \\
				\ding{172} &\yes& &59.3&59.3&59.2 &49.7 &61.6 \\
				\ding{173} &&\yes &55.4 &55.3  &55.6 &51.3 &64.5   \\
				\ding{174} &\yes&\yes&58.3 &58.6 &58.0 &53.9 &67.0\\
				\hline
	\end{tabular}}}
	\vspace{-2mm}
	\caption{Ablation studies on training volume and tunable parameters.
		ET: expending the training data.
		UFP: unfreeze all the parameters tunably.
		Results are reported in percentage (\%).
	}
	\vspace{-0mm}
	\label{tab:expand_data_parameters}
\end{table}

\textbf{Visualization.}
To explore how the learned modality-complementary prompts work in ViPT, we visualize several representative response maps and the corresponding t-SNE maps in Figure~\ref{fig:tsne}.
As can be seen, with auxiliary-modal prompts, ViPT has more unambiguous and discriminative responses when facing some complex scenarios, such as illumination variation and background cluster. t-SNE maps also verify that learned prompts help the foundation tracker distinguish the objects from the background due to the more separable distribution boundaries. 
As a result, it leads to more accurate and robust object localization overall.

\section{Conclusion}
In this work, we present ViPT, a new parameter-efficient vision tuning framework by introducing prompt-learning ideology to multi-modal tracking.
The core idea of ViPT is to maximally exploit the pre-trained foundation model which is trained at scale, and excavate the relevance of foundation and downstream tracking, and the complementarity of multiple modalities. Extensive experiments on multiple downstream tasks demonstrate the effectiveness and generalization of ViPT.
We expect this work can attract more attention to prompt learning for multi-modal tracking.

\noindent\textbf{Limitations.} 
Despite achieving a unified architecture for multi-modal tracking tasks, ViPT mainly focuses on visual prompting.
However, some vision-language tasks are emerging.
ViPT may have the potential to be extended to more tracking tasks like vision-language tracking. 
Moreover, our method still needs to train separately for the different multi-modal tasks. 
In the future, we are interested in joint training a general model for multiple modalities.
~\\

\noindent\textbf{Acknowledgements.}
This work was supported in part by the National Key R\&D Program of China under Grant No. 2018AAA0102001, the National Natural Science Foundation of China under Grant Nos. 62293542, U1903215, and the Fundamental Research Funds for the Central Universities under Grant No. DUT22ZD210.

\clearpage
{\small
	\bibliographystyle{ieee_fullname}
	\bibliography{egbib}

\begin{thebibliography}{10}\itemsep=-1pt

\bibitem{bahng2022visual}
Hyojin Bahng, Ali Jahanian, Swami Sankaranarayanan, and Phillip Isola.
\newblock Exploring visual prompts for adapting large-scale models.
\newblock {\em arXiv preprint arXiv:2203.17274}, 1(3):4, 2022.

\bibitem{siamesefc}
Luca Bertinetto, Jack Valmadre, Jo{\~a}o~F Henriques, Andrea Vedaldi, and
  Philip H~S Torr.
\newblock Fully-convolutional siamese networks for object tracking.
\newblock In {\em ECCVW}, 2016.

\bibitem{dimp}
Goutam Bhat, Martin Danelljan, Luc~Van Gool, and Radu Timofte.
\newblock Learning discriminative model prediction for tracking.
\newblock In {\em ICCV}, 2019.

\bibitem{adaptformer}
Shoufa Chen, Chongjian Ge, Zhan Tong, Jiangliu Wang, Yibing Song, Jue Wang, and
  Ping Luo.
\newblock Adaptformer: Adapting vision transformers for scalable visual
  recognition.
\newblock {\em NeurIPS}, 2022.

\bibitem{transt}
Xin Chen, Bin Yan, Jiawen Zhu, Dong Wang, Xiaoyun Yang, and Huchuan Lu.
\newblock Transformer tracking.
\newblock In {\em CVPR}, 2021.

\bibitem{conaire2008thermo}
Ciar{\'a}n~{\'O} Conaire, Noel~E O’Connor, and Alan Smeaton.
\newblock Thermo-visual feature fusion for object tracking using multiple
  spatiogram trackers.
\newblock {\em Machine Vision and Applications}, 19(5):483--494, 2008.

\bibitem{atom}
Martin Danelljan, Goutam Bhat, Fahad~Shahbaz Khan, and Michael Felsberg.
\newblock {ATOM: Accurate} tracking by overlap maximization.
\newblock In {\em CVPR}, 2019.

\bibitem{vit}
Alexey Dosovitskiy, Lucas Beyer, Alexander Kolesnikov, Dirk Weissenborn,
  Xiaohua Zhai, Thomas Unterthiner, Mostafa Dehghani, Matthias Minderer, Georg
  Heigold, Sylvain Gelly, Jakob Uszkoreit, and Neil Houlsby.
\newblock An image is worth 16x16 words: Transformers for image recognition at
  scale.
\newblock {\em ICLR}, 2021.

\bibitem{lasot}
Heng Fan, Liting Lin, Fan Yang, Peng Chu, Ge Deng, Sijia Yu, Hexin Bai, Yong
  Xu, Chunyuan Liao, and Haibin Ling.
\newblock {LaSOT}: A high-quality benchmark for large-scale single object
  tracking.
\newblock In {\em CVPR}, 2019.

\bibitem{dafnet}
Yuan Gao, Chenglong Li, Yabin Zhu, Jin Tang, Tao He, and Futian Wang.
\newblock Deep adaptive fusion network for high performance {RGBT} tracking.
\newblock In {\em ICCVW}, pages 0--0, 2019.

\bibitem{xavier}
Xavier Glorot and Yoshua Bengio.
\newblock Understanding the difficulty of training deep feedforward neural
  networks.
\newblock In {\em ICAIS}, 2010.

\bibitem{resnet}
Kaiming He, Xiangyu Zhang, Shaoqing Ren, and Jian Sun.
\newblock Deep residual learning for image recognition.
\newblock In {\em CVPR}, 2016.

\bibitem{got10k}
Lianghua Huang, Xin Zhao, and Kaiqi Huang.
\newblock Got-10k: A large high-diversity benchmark for generic object tracking
  in the wild.
\newblock {\em TPAMI}, 2019.

\bibitem{vpt}
Menglin Jia, Luming Tang, Bor-Chun Chen, Claire Cardie, Serge Belongie, Bharath
  Hariharan, and Ser-Nam Lim.
\newblock Visual prompt tuning.
\newblock In {\em ECCV}, 2022.

\bibitem{convpass}
Shibo Jie and Zhi-Hong Deng.
\newblock Convolutional bypasses are better vision transformer adapters.
\newblock {\em arXiv preprint arXiv:2207.07039}, 2022.

\bibitem{vot22}
Matej Kristan, Ale{\v{s}} Leonardis, Ji{\v{r}}{\'\i} Matas, Michael Felsberg,
  Roman Pflugfelder, Joni-Kristian K{\"a}m{\"a}r{\"a}inen, Hyung~Jin Chang,
  Martin Danelljan, Luka~{\v{C}}ehovin Zajc, Alan Luke{\v{z}}i{\v{c}}, et~al.
\newblock The tenth visual object tracking vot2022 challenge results.
\newblock In {\em ECCVW}, pages 431--460. Springer, 2023.

\bibitem{vot20}
Matej Kristan, Ale{\v{s}} Leonardis, Ji{\v{r}}{\'\i} Matas, Michael Felsberg,
  Roman Pflugfelder, Joni-Kristian K{\"a}m{\"a}r{\"a}inen, Martin Danelljan,
  Luka~{\v{C}}ehovin Zajc, Alan Luke{\v{z}}i{\v{c}}, Ondrej Drbohlav, et~al.
\newblock The eighth visual object tracking vot2020 challenge results.
\newblock In {\em ECCVW}, pages 547--601. Springer, 2020.

\bibitem{vot19}
Matej Kristan, Jiri Matas, Ales Leonardis, Michael Felsberg, Roman Pflugfelder,
  Joni-Kristian Kamarainen, Luka {\v{C}}ehovin~Zajc, Ondrej Drbohlav, Alan
  Lukezic, Amanda Berg, et~al.
\newblock The seventh visual object tracking vot2019 challenge results.
\newblock In {\em ICCVW}, pages 0--0, 2019.

\bibitem{vot21}
Matej Kristan, Ji{\v{r}}{\'\i} Matas, Ale{\v{s}} Leonardis, Michael Felsberg,
  Roman Pflugfelder, Joni-Kristian K{\"a}m{\"a}r{\"a}inen, Hyung~Jin Chang,
  Martin Danelljan, Luka Cehovin, Alan Luke{\v{z}}i{\v{c}}, et~al.
\newblock The ninth visual object tracking vot2021 challenge results.
\newblock In {\em ICCVW}, pages 2711--2738, 2021.

\bibitem{lester2021power}
Brian Lester, Rami Al-Rfou, and Noah Constant.
\newblock The power of scale for parameter-efficient prompt tuning.
\newblock {\em EMNLP}, 2021.

\bibitem{siamrpnplusplus}
Bo Li, Wei Wu, Qiang Wang, Fangyi Zhang, Junliang Xing, and Junjie Yan.
\newblock {SiamRPN++}: {Evolution} of siamese visual tracking with very deep
  networks.
\newblock In {\em CVPR}, 2019.

\bibitem{siameserpn}
Bo Li, Junjie Yan, Wei Wu, Zheng Zhu, and Xiaolin Hu.
\newblock High performance visual tracking with siamese region proposal
  network.
\newblock In {\em CVPR}, 2018.

\bibitem{rgbt234}
Chenglong Li, Xinyan Liang, Yijuan Lu, Nan Zhao, and Jin Tang.
\newblock {RGB-T} object tracking: Benchmark and baseline.
\newblock {\em Pattern Recognition}, 96:106977, 2019.

\bibitem{cat}
Chenglong Li, Lei Liu, Andong Lu, Qing Ji, and Jin Tang.
\newblock Challenge-aware {RGBT} tracking.
\newblock In {\em ECCV}, pages 222--237. Springer, 2020.

\bibitem{lasher}
Chenglong Li, Wanlin Xue, Yaqing Jia, Zhichen Qu, Bin Luo, Jin Tang, and Dengdi
  Sun.
\newblock Lasher: A large-scale high-diversity benchmark for {RGBT} tracking.
\newblock {\em IEEE Transactions on Image Processing}, 31:392--404, 2021.

\bibitem{sgt}
Chenglong Li, Nan Zhao, Yijuan Lu, Chengli Zhu, and Jin Tang.
\newblock Weighted sparse representation regularized graph learning for {RGB-T}
  object tracking.
\newblock In {\em ACMMM}, pages 1856--1864, 2017.

\bibitem{coco}
Tsung-Yi Lin, Michael Maire, Serge~J. Belongie, Lubomir~D. Bourdev, Ross~B.
  Girshick, James Hays, Pietro Perona, Deva Ramanan, Piotr Doll{\'a}r, and
  C.~Lawrence Zitnick.
\newblock {Microsoft COCO}: Common objects in context.
\newblock In {\em ECCV}, 2014.

\bibitem{liu2016combined}
Hongjie Liu, Diederik~Paul Moeys, Gautham Das, Daniel Neil, Shih-Chii Liu, and
  Tobi Delbr{\"u}ck.
\newblock Combined frame-and event-based detection and tracking.
\newblock In {\em ISCAS}, pages 2511--2514. IEEE, 2016.

\bibitem{spm}
Lingbo Liu, Bruce~XB Yu, Jianlong Chang, Qi Tian, and Chang-Wen Chen.
\newblock Prompt-matched semantic segmentation.
\newblock {\em arXiv preprint arXiv:2208.10159}, 2022.

\bibitem{liu2021pre}
Pengfei Liu, Weizhe Yuan, Jinlan Fu, Zhengbao Jiang, Hiroaki Hayashi, and
  Graham Neubig.
\newblock Pre-train, prompt, and predict: A systematic survey of prompting
  methods in natural language processing.
\newblock {\em arXiv preprint arXiv:2107.13586}, 2021.

\bibitem{ca3dms}
Ye Liu, Xiao-Yuan Jing, Jianhui Nie, Hao Gao, Jun Liu, and Guo-Ping Jiang.
\newblock Context-aware three-dimensional mean-shift with occlusion handling
  for robust object tracking in {RGB-D} videos.
\newblock {\em IEEE Transactions on Multimedia}, 21(3):664--677, 2018.

\bibitem{adamw}
Ilya Loshchilov and Frank Hutter.
\newblock Decoupled weight decay regularization.
\newblock In {\em ICLR}, 2018.

\bibitem{luo2019thermal}
Chengwei Luo, Bin Sun, Ke Yang, Taoran Lu, and Wei-Chang Yeh.
\newblock Thermal infrared and visible sequences fusion tracking based on a
  hybrid tracking framework with adaptive weighting scheme.
\newblock {\em Infrared Physics \& Technology}, 99:265--276, 2019.

\bibitem{miangoleh2021boosting}
S~Mahdi~H Miangoleh, Sebastian Dille, Long Mai, Sylvain Paris, and Yagiz Aksoy.
\newblock Boosting monocular depth estimation models to high-resolution via
  content-adaptive multi-resolution merging.
\newblock In {\em CVPR}, pages 9685--9694, 2021.

\bibitem{mueller2017real}
Franziska Mueller, Dushyant Mehta, Oleksandr Sotnychenko, Srinath Sridhar, Dan
  Casas, and Christian Theobalt.
\newblock Real-time hand tracking under occlusion from an egocentric {RGB-D}
  sensor.
\newblock In {\em ICCV}, pages 1154--1163, 2017.

\bibitem{trackingnet}
Matthias Muller, Adel Bibi, Silvio Giancola, Salman Alsubaihi, and Bernard
  Ghanem.
\newblock Tracking{N}et: A large-scale dataset and benchmark for object
  tracking in the wild.
\newblock In {\em ECCV}, 2018.

\bibitem{dal}
Yanlin Qian, Song Yan, Alan Luke{\v{z}}i{\v{c}}, Matej Kristan, Joni-Kristian
  K{\"a}m{\"a}r{\"a}inen, and Ji{\v{r}}{\'\i} Matas.
\newblock Dal: A deep depth-aware long-term tracker.
\newblock In {\em ICPR}, pages 7825--7832. IEEE, 2021.

\bibitem{clip}
Alec Radford, Jong~Wook Kim, Chris Hallacy, Aditya Ramesh, Gabriel Goh,
  Sandhini Agarwal, Girish Sastry, Amanda Askell, Pamela Mishkin, Jack Clark,
  et~al.
\newblock Learning transferable visual models from natural language
  supervision.
\newblock In {\em ICML}, pages 8748--8763. PMLR, 2021.

\bibitem{imagenet}
Olga Russakovsky, Jia Deng, Hao Su, Jonathan Krause, Sanjeev Satheesh, Sean Ma,
  Zhiheng Huang, Andrej Karpathy, Aditya Khosla, and Michael Bernstein.
\newblock {ImageNet} {Large} scale visual recognition challenge.
\newblock {\em IJCV}, 2015.

\bibitem{tsne}
Laurens Van~der Maaten and Geoffrey Hinton.
\newblock Visualizing data using t-{SNE}.
\newblock {\em Journal of machine learning research}, 9(11), 2008.

\bibitem{attention_is_all}
Ashish Vaswani, Noam Shazeer, Niki Parmar, Jakob Uszkoreit, Llion Jones,
  Aidan~N Gomez, Lukasz Kaiser, and Illia Polosukhin.
\newblock Attention is all you need.
\newblock In {\em NIPS}, 2017.

\bibitem{cmpp}
Chaoqun Wang, Chunyan Xu, Zhen Cui, Ling Zhou, Tong Zhang, Xiaoya Zhang, and
  Jian Yang.
\newblock Cross-modal pattern-propagation for {RGB-T} tracking.
\newblock In {\em CVPR}, pages 7064--7073, 2020.

\bibitem{visevent}
Xiao Wang, Jianing Li, Lin Zhu, Zhipeng Zhang, Zhe Chen, Xin Li, Yaowei Wang,
  Yonghong Tian, and Feng Wu.
\newblock Visevent: Reliable object tracking via collaboration of frame and
  event flows.
\newblock {\em arXiv preprint arXiv:2108.05015}, 2021.

\bibitem{apfnet}
Yun Xiao, Mengmeng Yang, Chenglong Li, Lei Liu, and Jin Tang.
\newblock Attribute-based progressive fusion network for {RGBT} tracking.
\newblock In {\em AAAI}, 2022.

\bibitem{youtube-vos}
Ning Xu, Linjie Yang, Yuchen Fan, Jianchao Yang, Dingcheng Yue, Yuchen Liang,
  Brian Price, Scott Cohen, and Thomas Huang.
\newblock Youtube-vos: Sequence-to-sequence video object segmentation.
\newblock In {\em ECCV}, pages 585--601, 2018.

\bibitem{stark}
Bin Yan, Houwen Peng, Jianlong Fu, Dong Wang, and Huchuan Lu.
\newblock Learning spatio-temporal transformer for visual tracking.
\newblock In {\em ICCV}, 2021.

\bibitem{depthtrack}
Song Yan, Jinyu Yang, Jani K{\"a}pyl{\"a}, Feng Zheng, Ale{\v{s}} Leonardis,
  and Joni-Kristian K{\"a}m{\"a}r{\"a}inen.
\newblock Depthtrack: Unveiling the power of {RGBD} tracking.
\newblock In {\em ICCV}, pages 10725--10733, 2021.

\bibitem{protrack}
Jinyu Yang, Zhe Li, Feng Zheng, Ales Leonardis, and Jingkuan Song.
\newblock Prompting for multi-modal tracking.
\newblock In {\em ACMMM}, pages 3492--3500, 2022.

\bibitem{ostrack}
Botao Ye, Hong Chang, Bingpeng Ma, Shiguang Shan, and Xilin Chen.
\newblock Joint feature learning and relation modeling for tracking: A
  one-stream framework.
\newblock In {\em ECCV}, pages 341--357. Springer, 2022.

\bibitem{macnet}
Hui Zhang, Lei Zhang, Li Zhuo, and Jing Zhang.
\newblock Object tracking in {RGB-T} videos using modal-aware attention network
  and competitive learning.
\newblock {\em Sensors}, 20(2):393, 2020.

\bibitem{stnet}
Jiqing Zhang, Bo Dong, Haiwei Zhang, Jianchuan Ding, Felix Heide, Baocai Yin,
  and Xin Yang.
\newblock Spiking transformers for event-based single object tracking.
\newblock In {\em CVPR}, pages 8801--8810, 2022.

\bibitem{fenet}
Jiqing Zhang, Xin Yang, Yingkai Fu, Xiaopeng Wei, Baocai Yin, and Bo Dong.
\newblock Object tracking by jointly exploiting frame and event domain.
\newblock In {\em ICCV}, pages 13043--13052, 2021.

\bibitem{mfdimp}
Lichao Zhang, Martin Danelljan, Abel Gonzalez-Garcia, Joost van~de Weijer, and
  Fahad Shahbaz~Khan.
\newblock Multi-modal fusion for end-to-end {RGB-T} tracking.
\newblock In {\em ICCVW}, pages 0--0, 2019.

\bibitem{zhang2021learning}
Pengyu Zhang, Dong Wang, Huchuan Lu, and Xiaoyun Yang.
\newblock Learning adaptive attribute-driven representation for real-time
  {RGB-T} tracking.
\newblock {\em IJCV}, 129(9):2714--2729, 2021.

\bibitem{jmmac}
Pengyu Zhang, Jie Zhao, Chunjuan Bo, Dong Wang, Huchuan Lu, and Xiaoyun Yang.
\newblock Jointly modeling motion and appearance cues for robust {RGB-T}
  tracking.
\newblock {\em IEEE Transactions on Image Processing}, 30:3335--3347, 2021.

\bibitem{vtuav}
Pengyu Zhang, Jie Zhao, Dong Wang, Huchuan Lu, and Xiang Ruan.
\newblock Visible-thermal uav tracking: A large-scale benchmark and new
  baseline.
\newblock In {\em CVPR}, pages 8886--8895, 2022.

\bibitem{siamcda}
Tianlu Zhang, Xueru Liu, Qiang Zhang, and Jungong Han.
\newblock Siamcda: Complementarity-and distractor-aware {RGB-T} tracking based
  on siamese network.
\newblock {\em TCSVT}, 32(3):1403--1417, 2021.

\bibitem{fc2019fusion}
Xingchen Zhang, Ping Ye, Dan Qiao, Junhao Zhao, Shengyun Peng, and Gang Xiao.
\newblock Object fusion tracking based on visible and infrared images using
  fully convolutional siamese networks.
\newblock In {\em FUSION}, pages 1--8. IEEE, 2019.

\bibitem{doprompt}
Zangwei Zheng, Xiangyu Yue, Kai Wang, and Yang You.
\newblock Prompt vision transformer for domain generalization.
\newblock {\em arXiv preprint arXiv:2208.08914}, 2022.

\bibitem{rgbd1k}
Xue-Feng Zhu, Tianyang Xu, Zhangyong Tang, Zucheng Wu, Haodong Liu, Xiao Yang,
  Xiao-Jun Wu, and Josef Kittler.
\newblock {RGBD1K}: A large-scale dataset and benchmark for {RGB-D} object
  tracking.
\newblock {\em AAAI}, 2023.

\bibitem{dapnet}
Yabin Zhu, Chenglong Li, Bin Luo, Jin Tang, and Xiao Wang.
\newblock Dense feature aggregation and pruning for {RGBT} tracking.
\newblock In {\em ACMMM}, pages 465--472, 2019.

\bibitem{fanet}
Yabin Zhu, Chenglong Li, Jin Tang, and Bin Luo.
\newblock Quality-aware feature aggregation network for robust {RGBT} tracking.
\newblock {\em IEEE Transactions on Intelligent Vehicles}, 6(1):121--130, 2020.

\end{thebibliography}
}

\end{document}